\documentclass{article}

\usepackage[dblblindworkshop,final]{NeurIPS-2025-Multi-Turn-Interactions-in-Large-Language-Models/neurips_2025}
\makeatletter
\renewcommand{\footnoterule}{%
  \kern -2pt
  \hrule \@width \columnwidth height 0.4pt
  \kern 4pt
}
\makeatother

\usepackage{wrapfig}
\usepackage[utf8]{inputenc} 
\usepackage[T1]{fontenc}    
\usepackage{hyperref}       
\DeclareUnicodeCharacter{03C4}{\ensuremath{\tau}}
\usepackage{url}            
\usepackage{booktabs}       
\usepackage{amsfonts}       
\usepackage{nicefrac}       
\usepackage{microtype}      
\usepackage{xcolor}         
\usepackage{graphicx}
\usepackage[ruled,vlined]{algorithm2e}
\usepackage{subcaption}
\usepackage{amsmath}
\usepackage[many]{tcolorbox}
\NewTColorBox[auto counter]{NewBox}{ s O{!htbp} m m }{%
  floatplacement={#2},                        
  IfBooleanTF={#1}{float*,width=\textwidth}{float},  
  fontupper=\small, 
  title={Prompt \thetcbcounter: #4},
  label={#3}                                  
}

\title{Evaluating Multi-Turn Bargain Skills in LLM-Based Seller Agents}
\workshoptitle{Multi-Turn Interactions in LLMs, under review}

\author{%
\begin{minipage}{0.99\textwidth}\centering
\fontsize{9pt}{12pt}\selectfont
{\bfseries
Issue~Yishu~Wang$^{1,*}$,\;
Kakam~Chong$^{2,*}$,\;
Xiaofeng~Wang$^{3,*}$,\;
Xu~Yan$^{4}$,\; Dexin~Kong$^{4}$,\\
Chen~Ju$^{4}$,\; Ming~Chen$^{4}$,\;
Shuai~Xiao$^{4}$,\; Shuguang~Han$^{4}$,\; Junfeng~Chen$^{4}$%
}\\[2pt]
{\bfseries
$^{1}$Johns Hopkins University\,
$^{2}$Tsinghua University\,
$^{3}$Shanghai Jiao Tong University\,
$^{4}$Alibaba Group%
}\\[2pt]
\normalfont\mdseries
\texttt{ywan1000@jhu.edu, zhuangjx23@mails.tsinghua.edu.cn, banyedy@sjtu.edu.cn}\\[-0.2ex]
\texttt{\{wuyong.yx, kongdexin.kdx, juchen.ju, xingke.cm, shuai.xsh, shuguang.sh, ufeng.cjf\}@alibaba-inc.com}
\end{minipage}
}

\makeatletter
\@ifpackageloaded{lineno}{\AtBeginDocument{\nolinenumbers}}{}
\makeatother

\begin{document}

\maketitle
\vspace{-0.5\baselineskip}

\begingroup
\renewcommand\thefootnote{\fnsymbol{footnote}}
\footnotetext[1]{Equal contribution.}
\footnotetext[2]{Work conducted during internships at Alibaba Group.}
\endgroup

\begin{abstract}
In online second-hand marketplaces, multi-turn bargaining is a crucial part of seller-buyer interactions. Large Language Models (LLMs) can act as \emph{seller agents}, negotiating with buyers on behalf of sellers under given business constraints. A critical ability for such agents is to track and accurately interpret cumulative buyer intents across long negotiations, which directly impacts bargaining effectiveness.
We introduce a multi-turn evaluation framework for measuring the bargaining ability of seller agents in e-commerce dialogues. The framework tests whether an agent can extract and track buyer intents.
Our contributions are: (1) a large-scale e-commerce bargaining benchmark spanning 622 categories, 9{,}892 products, and 3{,}014 tasks; (2) a turn-level evaluation framework grounded in Theory of Mind (ToM) with annotated buyer intents, moving beyond outcome-only metrics; and (3) an automated pipeline that extracts reliable intent from massive dialogue data.

\end{abstract}

\section{Introduction}

\begin{wrapfigure}{rt}{0.35\textwidth}
\vspace{-13pt}
    \centering
    \includegraphics[width=1\linewidth]{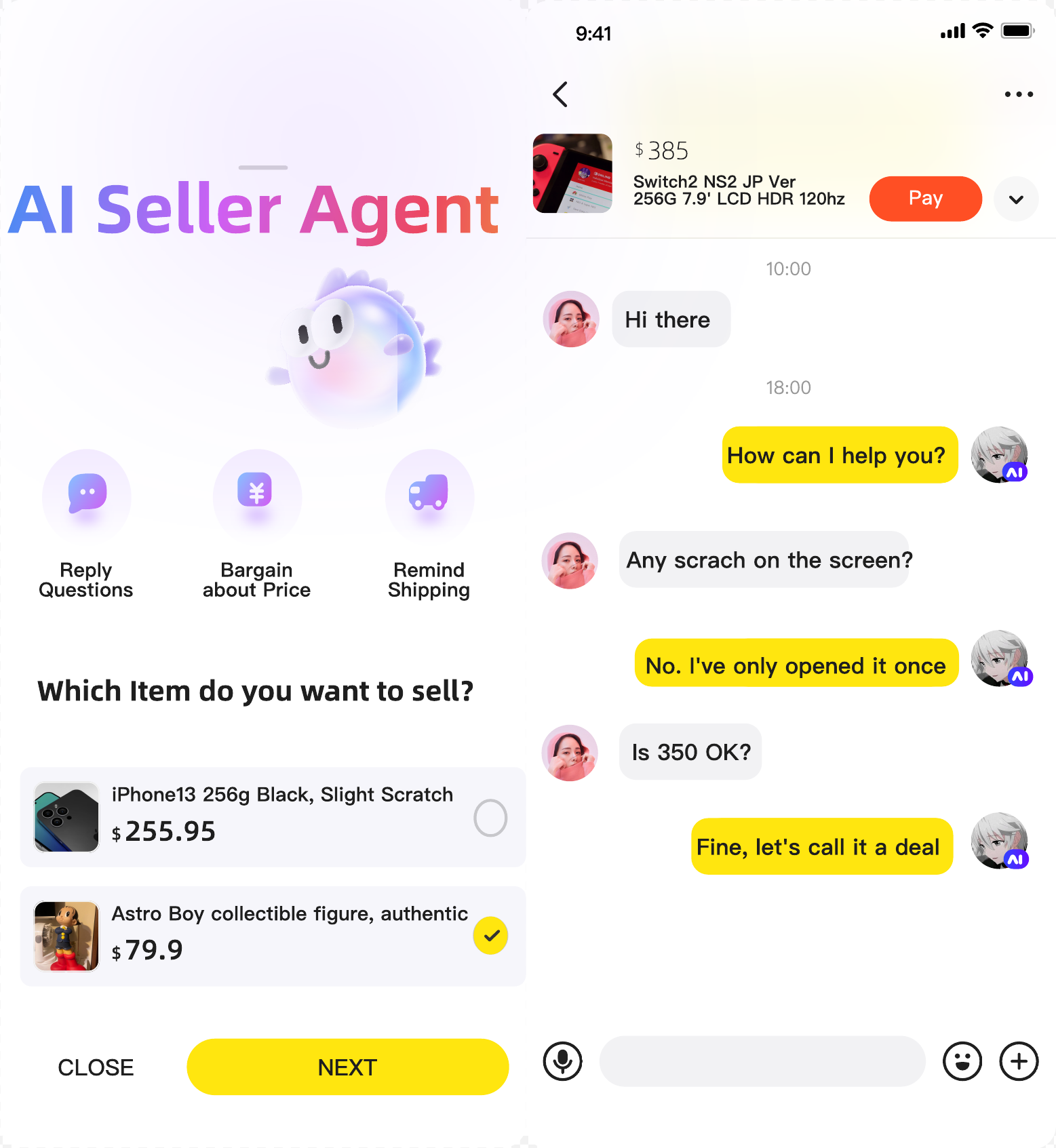}
    \vspace{-0.25in}
    \caption{\label{fig:framework-intro-overview}Left: Human sellers select items and delegate them to the AI seller agent. Right: Human buyers interact with the AI agent to ask questions and negotiate prices}
\vspace{-0.2in}
\end{wrapfigure}

Bargaining is a fundamental social intelligence skill, widely applied in e-commerce, diplomacy, and labor negotiations\cite{he_decoupling_2018}. It requires the interpretation of scenario-specific information, reasoning about counterpart goals and constraints \cite{davidson_evaluating_2024}, and planning actions toward mutually beneficial agreements.  

Current LLMs struggle with bargaining because they misread adversarial intents and rely on shallow pattern-matching instead of strategic reasoning. Effective bargaining also requires accurately tracking counterpart intents across multiple turns, integrating prior context \cite{dexin_fishbargain_2025}, and applying such understanding under domain-specific constraints, a setting where existing models remain fragile.  

Prior benchmarks either ignore real-world constraints or score only final deals \cite{xia_measuring_2024}, missing the intermediate reasoning that shapes negotiation success. We introduce the seller-agent setting, where agents negotiate on behalf of sellers under explicit business rules, to isolate and measure turn-level intent understanding in a controlled yet realistic environment.  

Our work makes three main contributions. First, we present a bargaining benchmark on e-commerce which is substantially larger and more challenging than prior datasets. It  cover 622 categories, 9{,}892 product listings, and 3{,}014 tasks. Second, we introduce a turn-level task design grounded in Theory of Mind (ToM), which provides ground-truth buyer intents and shifts evaluation from outcome-only metrics to the intermediate reasoning process. Third, we develop an automated pipeline for extracting high-quality intent from large-scale dialogues, enabling scalable and reproducible benchmarking of bargaining agents.

\section{Related Work}

Early bargaining datasets such as DealOrNoDeal~\cite{lewis_deal_2017} and CraigslistBargain~\cite{he_decoupling_2018} established text-based negotiation protocols, later extended to more interactive and applied scenarios, e.g., FishBargain~\cite{dexin_fishbargain_2025}. Intent recognition in dialogue has been widely studied in Dialogue State Tracking~\cite{budzianowski_multiwoz_2018}, yet bargaining often involves implicit intents, which remain underexplored~\cite{guan_evaluating_2025,chan_negotiationtom_2024}. Tool-augmented dialogue benchmarks, such as $\tau$-Bench~\cite{yao_-bench_2024} and ToolACE~\cite{liu_toolace_2025}, emphasize robustness and correctness under domain constraints.

Our work differs from prior studies by shifting the focus from outcome-based negotiation metrics to turn-level buyer intent recognition, grounded in an \emph{intent–action–tool} hierarchy and emphasizing intent alignment in bargaining. More related work is provided in Appendix~\ref{appendix:related}.

\section{Methodology}

\begin{figure}[tp]
\vspace{-0.2in}
    \centering
    \includegraphics[width=0.85\linewidth]{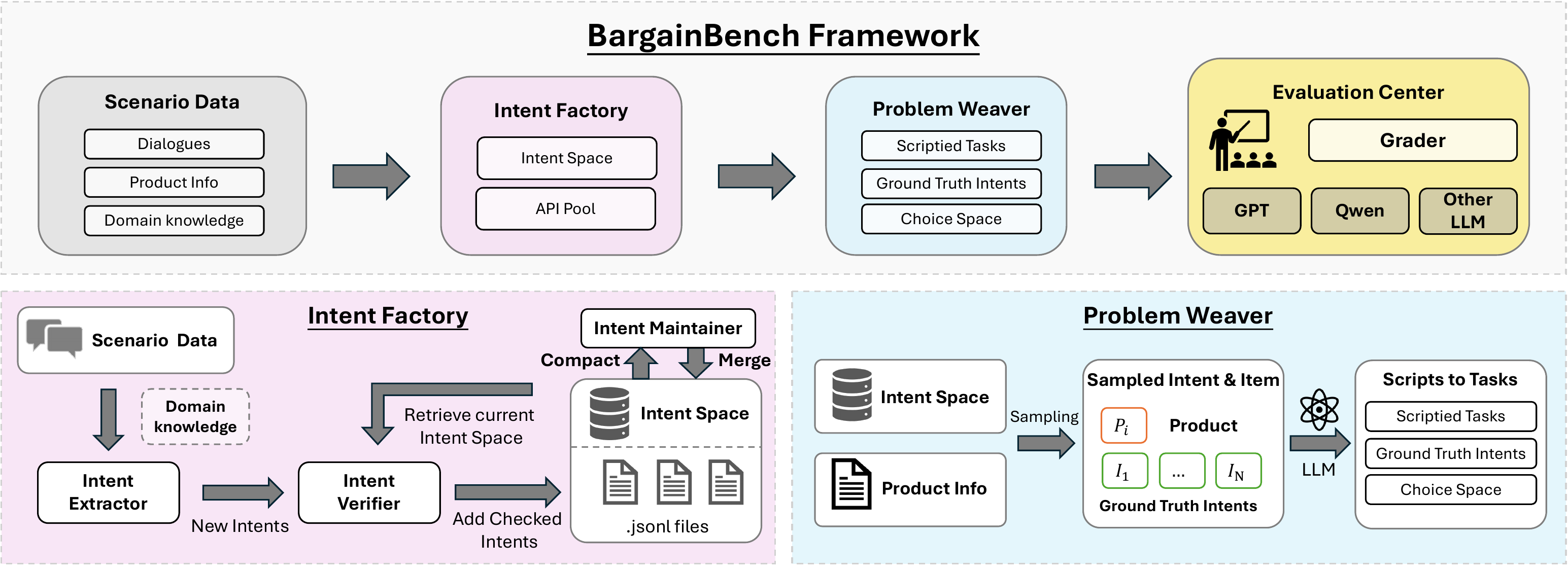}
    \caption{BargainBench framework: \textbf{Intent Factory} extracts an intent space, \textbf{Problem Weaver} generates scripted dialogues, and \textbf{Evaluation Center} scores LLM performance.}

    \label{fig:framework}
\vspace{-0.2in}
\end{figure}

The framework of Bargainbench is shown in Figure~\ref{fig:framework}. The Bargainbench consists of three main components: \textbf{Intent Factory}, \textbf{Problem Weaver} and \textbf{Evaluation Center}. Advantage discussion regarding the framework is provided in appendix~\ref{appendix:framework-advantage}. We decompose negotiation behavior into a three-level hierarchy: \emph{intent} (high-level goal), \emph{action} (mid-level negotiation move), and \emph{tool}—the most atomic, directly executable intention that can be unambiguously checked in a single turn. This tree structure disentangles complex intents from raw dialogues and allows fine-grained diagnosis of model strengths and weaknesses.

\subsection{Intent Factory}

The \textbf{Intent Factory} distils raw dialogues, product data, and domain knowledge into the three-level \emph{intent–action–tool} hierarchy and a compact API pool. A lightweight multi-agent pipeline (Extractor, Verifier, Maintainer) successively refine intent quality and remove duplicated entries; full roles and convergence analysis are provided in Appendix \ref{appendix:intent-factory-full}.



\subsection{Problem Weaver}

The \textbf{Problem Weaver} converts entries from the API Pool into concrete multi-turn bargaining scenarios using publicly available product metadata (e.g., title, description, price, category). Given product information, an intent pool, and predefined prompts, it samples a product–intent sequence, verifies plausibility, and prompts the LLM to generate turn-specific buyer queries grounded in product attributes. Each query is paired with a system prompt and ground truth intent, producing structured dialogue data with scripts, intent labels, and choice sets. Processing details and question sample is provided in appendix ~\ref{appendix:problem-weaver} and ~\ref{appendix:task-sample}

\subsection{Evaluation Center}

The \textbf{Evaluation Center} executes these scenarios on target LLMs and scores predictions against turn-level ground truth, with the maximum per-dialogue score equal to the number of annotated intents. The evaluation checks whether outputs meet format requirements, predicted intents exist in the intent space, and the predicted sequence matches the reference. At each turn, the model receives the dialogue so far, product information, and candidate intents, and must output buyer intents in order. The grader computes per-turn accuracy and aggregated scores for model comparison.



\section{Benchmark}
\paragraph{Task Formulation.}
The evaluation task is defined as follows. The model input consists of three components: (1) the \textbf{dialogue history}, i.e., the full multi-turn bargaining context up to the current turn, ensuring that no prior information is omitted; (2) the \textbf{product information}, which includes real-world item descriptions, hierarchical category metadata with four levels, and listing prices; and (3) the \textbf{intent choice space}, a set of 20 candidate options randomly sampled from the complete intent space. The model output is a prediction of the buyer’s true intent at each turn, selected from the choice space. The central challenge is whether the model can continuously track and correctly identify buyer intent throughout multi-turn interactions.

\paragraph{Data Overview.}
The detail of data is shown in Table~\ref{tab:dataset-stats}
More details can be found in appendix \ref{appendix:data-prep-details}



\begin{table}[ht]
\centering
\caption{Dataset Statistics}
\begin{tabular}{l r}
\toprule
\textbf{Statistic} & \textbf{Value} \\
\midrule
Total Items & 9,892 \\
Unique Level 1 Categories & 85 \\
Unique Level 2 Categories & 700 \\
Unique Level 3 Categories & 1,336 \\
Unique Level 4 Categories & 1,611 \\
\bottomrule
\end{tabular}
\label{tab:dataset-stats}
\end{table}


\paragraph{Metrics.}

We categorize predicted intents into four types.
(1) \textbf{Correct Intent (CI)}: predicted Intents that are in the choice space and exactly match the ground truth.
(2) \textbf{Mismatched Intent (MMI)}: predicted intents that are in the choice space but do not match the ground truth.
(3) \textbf{Missed Intent (MI)}: Intents that are in ground truth but not in predicted actions
(4) \textbf{Invalid Intent (II)}: actions that are not in choice space.

Metrics are defined as: 
(1) \textbf{Intent-Precision (IP)}: precision measures the proportion of correct predictions among all predictions:
    $\frac{\text{CI}}{\text{CI}+\text{MMI}+\text{II}}$.
(2) \textbf{Intent-Recall (IR)}: recall measures the proportion of ground truth actions correctly predicted: $\frac{\text{CI}}{\text{CI}+\text{MI}}$
(3) \textbf{Intent-F1}: F1 balances precision and recall: $F1(\text{IR},\text{IP})=\frac{2\cdot \text{IR}\cdot \text{IP}}{\text{IR} + \text{IP}}$
(4) \textbf{Failure rate}: this reflects the proportion of invalid predictions: $\frac{\text{II}}{\text{CI} + \text{MI} + \text{II}}$




\section{Results}

The main results are shown in Table~\ref{tab:main_result}.



\begin{table}[htp]
\centering
\caption{Main experimental results. All metric values in this table are reported as percentages (\%). Best and second results are in bold and underlined, respectively. Each column header “Turn-N” refers to tasks comprising N turns. }
\resizebox{\textwidth}{!}{%
    \begin{tabular}{lcccccccccccc}
    \toprule
    \multicolumn{1}{c}{} & \multicolumn{4}{c}{\textbf{Turn-2}} & \multicolumn{4}{c}{\textbf{Turn-3}} & \multicolumn{4}{c}{\textbf{Turn-4+}} \\
    \cmidrule(lr){2-5} \cmidrule(lr){6-9} \cmidrule(lr){10-13}
    \textbf{Model} & \textbf{Precision} $\uparrow$ & \textbf{Recall} $\uparrow$ & \textbf{Failurs} $\downarrow$ & \textbf{F1} $\uparrow$ & \textbf{Precision} $\uparrow$ & \textbf{Recall} $\uparrow$ & \textbf{Failure} $\downarrow$ & \textbf{F1} $\uparrow$ & \textbf{Precision} $\uparrow$ & \textbf{Recall} $\uparrow$ & \textbf{Failure} $\downarrow$ & \textbf{F1} $\uparrow$ \\
    \midrule
     gpt-5-chat-0807-global &       \textbf{52.55} &    44.72 &   \textbf{0.00} & \underline{48.32} &       \textbf{56.73} &    48.05 &   \textbf{0.00} & \underline{52.03} &       \textbf{51.31} &    42.19 &   \textbf{\textbf{0.00}} & 46.30 \\
 gemini-2.5-pro-06-17   &       44.24 &    \underline{48.91} &   9.13 & 46.46 &       48.31 &    \underline{53.02} &   8.88 & 50.56 &       45.57 &    \textbf{49.03} &   7.47 & \underline{47.24} \\
 gpt-41-0414-global     &       \underline{50.08} &    46.43 &   \underline{0.17} & 48.19 &       51.94 &    47.08 &   0.11 & 49.39 &       \underline{48.85} &    44.35 &   0.08 & 46.49 \\
 o3-0416-global         &       47.09 &    42.70 &   0.86 & 44.79 &       50.55 &    44.83 &   0.11 & 47.52 &       47.69 &    41.44 &   0.43 & 44.35 \\
 DeepSeek-V3-671B       &       20.59 &    23.76 &  59.89 & 22.06 &       24.69 &    28.85 &  52.13 & 26.61 &       19.91 &    22.40 &  56.61 & 21.08 \\
 kimi-k2                &       46.59 &    46.74 &   \textbf{\textbf{0.00}} & 46.67 &       46.55 &    47.27 &   \textbf{0.00} & 46.91 &       45.40 &    45.16 &   \underline{0.07} & 45.28 \\
 qwen2.5-72b-instruct   &       49.02 &    \textbf{50.62} &   1.20 & \textbf{49.81} &       \underline{53.77} &    \textbf{56.34} &   0.37 & \textbf{55.02} &       47.89 &    \textbf{49.78} &   0.29 & \textbf{48.81} \\
 qwen3-14b              &       44.60 &    45.50 &   \textbf{0.00} & 45.04 &       46.19 &    47.27 &   \underline{0.10} & 46.72 &       44.18 &    44.05 &   \textbf{\textbf{0.00}} & 44.11 \\
 qwen3-32b              &       48.50 &    47.83 &   1.10 & 48.16 &       52.84 &    51.66 &   2.69 & 52.24 &       48.28 &    45.83 &   3.06 & 47.02 \\
\hline
    \bottomrule
    \end{tabular}
} 

\label{tab:main_result}
\end{table}

\paragraph{Stable models show near-zero failure, while weaker ones collapse.}
Failure directly reflects the reliability of a model acting as a seller agent. GPT-5-chat and Kimi-K2 achieve almost zero failure across all turns, while Qwen2.5-72B-Instruct and Qwen-14B remain similarly stable with values below 1.2\%. In contrast, DeepSeek-V3-671B collapses with failure above 50\%, and Gemini fluctuates near 8\%, underscoring weaker robustness in multi-turn bargaining.

\paragraph{GPT-5 is the strongest performer, with Qwen competitive on F1.}
GPT-5-chat combines the highest precision (56.7\%) with perfect stability, yielding the most reliable overall performance. Qwen2.5-72B-Instruct achieves the best F1 balance (55.0\% at Turn-3), supported by strong recall, while Qwen-32B is close behind. Kimi-K2 remains extremely stable but less precise, and DeepSeek-V3-671B performs worst with F1 below 27\%.

\paragraph{Precision distinguishes strong models, while recall remains steady.}
Precision separates strong from weak systems more clearly than recall. GPT-5-chat leads with the highest precision, while Qwen2.5-72B-Instruct also performs well above 53\%. Recall varies less across systems, with Qwen2.5-72B-Instruct consistently leading, followed by Gemini. Overall, recall remains steady while precision determines consistency of intent prediction.

\paragraph{Additional turns improve understanding but amplify inconsistency.}
Most models show small gains at Turn-3 and mild declines at Turn-4+. For example, Qwen2.5-72B-Instruct rises from 49.8\% to 55.0\% before dropping to 48.8\%. This suggests additional turns can enhance task understanding, but longer dialogues mainly amplify inconsistency (precision loss) rather than coverage errors (recall remains steady).

\paragraph{Structured queries are easy, while ambiguous or rare intents remain weak.}
Accuracy differs sharply across intent types. Structured and semantically well-defined queries achieve the highest rates: product authenticity (\texttt{Inquire\_Product\_Authenticity} / \texttt{QueryProductAuthenticity}), terminology explanations (\texttt{ExplainTerm}), and policy lookups (\texttt{GetPriceAndNegotiationPolicy}) reach 83--87\%. By contrast, ambiguous or infrequent intents perform poorly: \texttt{PromoteLogisticsService} (4.17\%), \texttt{IgnoreInappropriateRequest} (15.69\%), \texttt{Tool\_Greet} (18.18\%), and \texttt{Business\_Cooperation} (4.17\%). These categories often overlap multiple domains or lie outside the core bargaining process, leading to insufficient training coverage and frequent confusion.

\begin{figure}[htbp]
\vspace{-0.1in}
  \centering
  \begin{subfigure}[t]{0.48\textwidth}
    \centering
    \includegraphics[width=\linewidth]{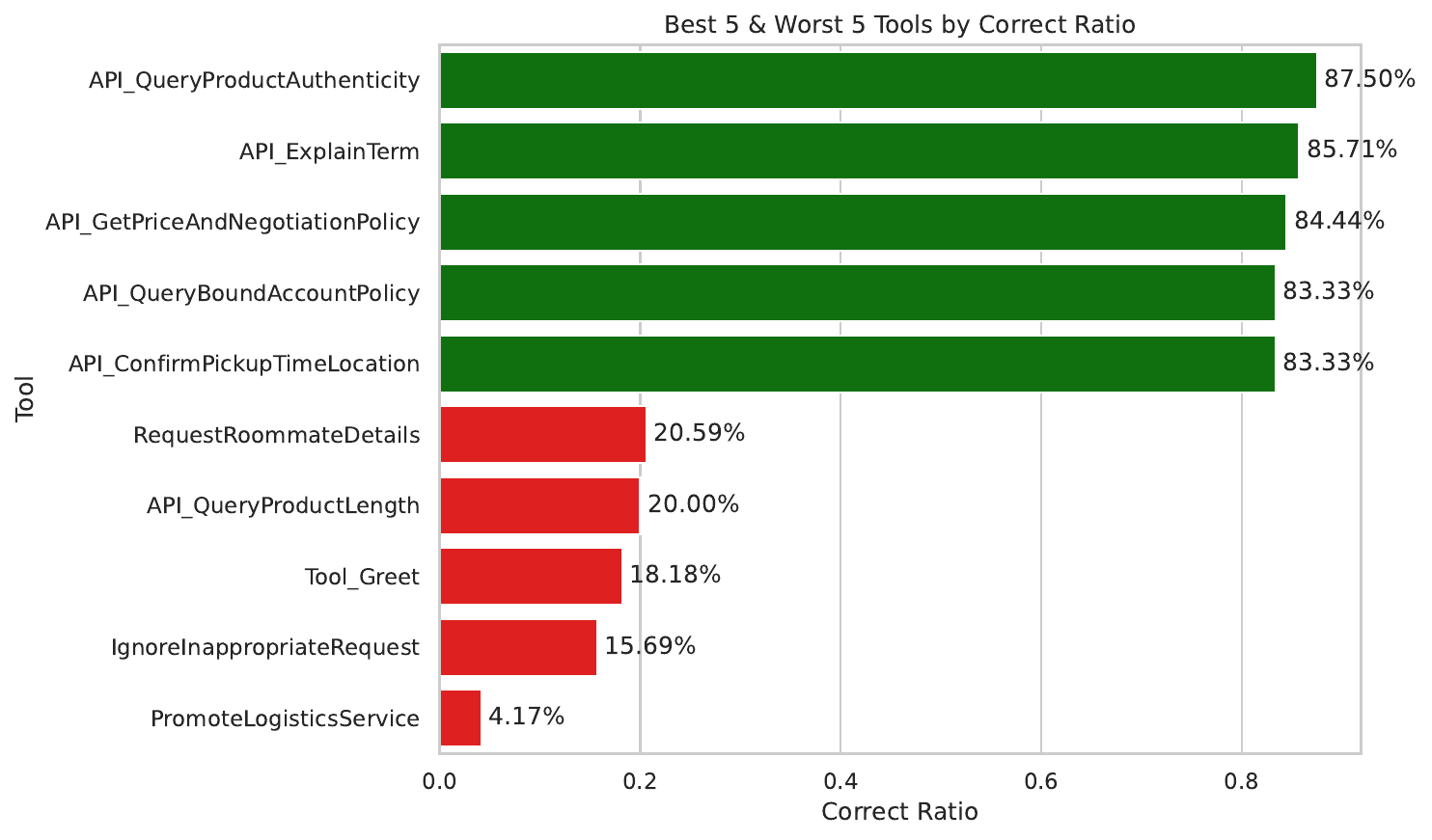}
    \caption{Top recognized tool and the worst one.}
    \label{fig:best_worst_tools}
  \end{subfigure}\hfill
  \begin{subfigure}[t]{0.48\textwidth}
    \centering
    \includegraphics[width=\linewidth]{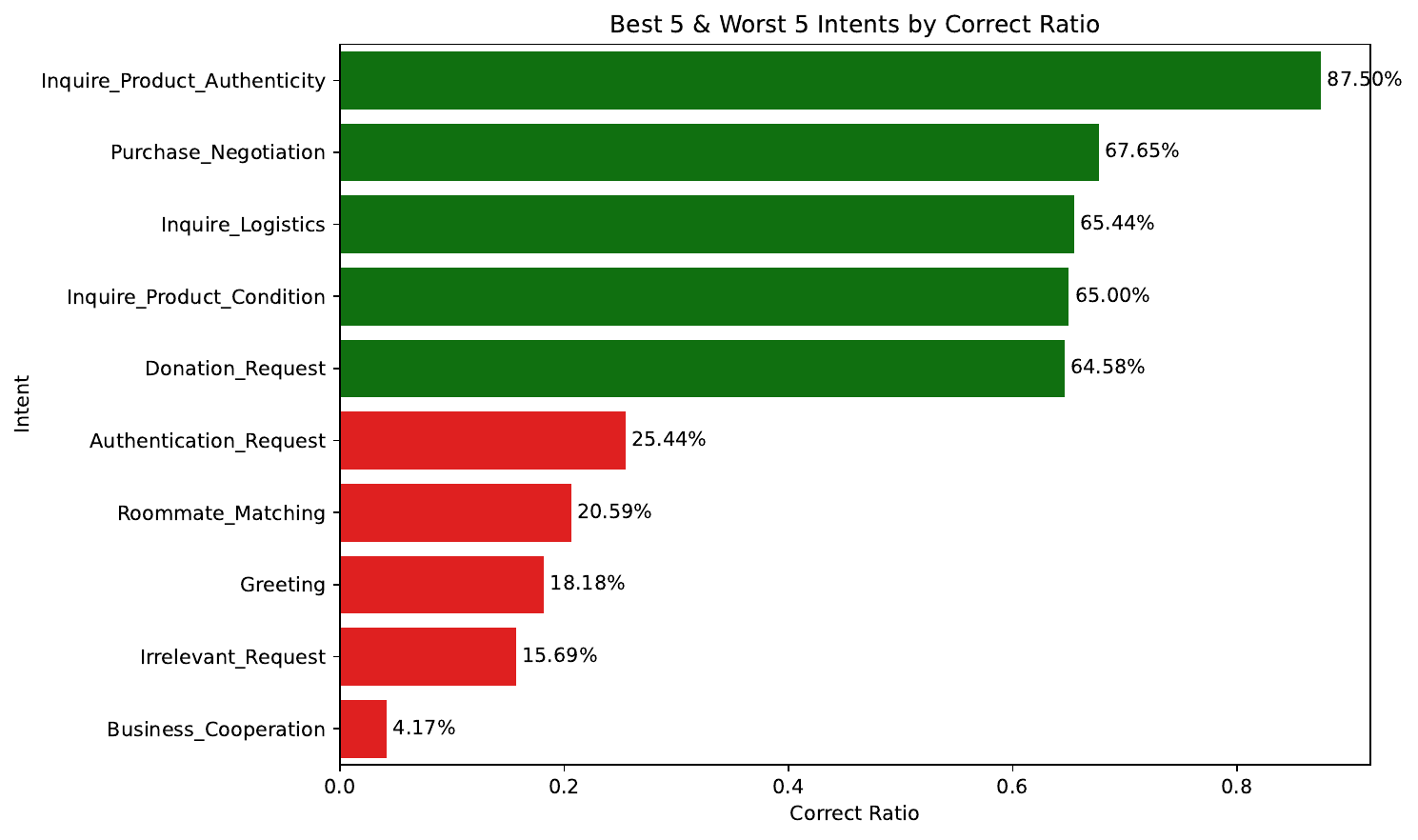}
    \caption{Performance Aggregated in intent level.}
    \label{fig:best_worst_intents}
  \end{subfigure}
  \caption{Tools vs. Intents distributions.}
  \label{fig:tool-rank} 
\end{figure}

\section{Conclusion}
We present a business-grounded multi-turn benchmark for evaluating seller agents, featuring intent recognition and verifiable evaluation. 
Future work includes extending evaluations to single-turn scenarios, and enriching the benchmark with \textit{special} and \textit{agent} settings—where \textit{special} cases assess the ability to identify and appropriately refuse responses beyond buyer-agent capabilities, and \textit{agent} cases involve three-party dialogues when seller return to conversation. We also expect to adapt our framework to other scenarios such as diplomacy, persuasion, and games.


\appendix
\section*{Appendix} 
\section{Intent Factory: Multi-Agent Pipeline \& Quality Metrics}
\label{appendix:intent-factory-full}
\addcontentsline{toc}{section}{Intent Factory: Multi-Agent Pipeline \& Quality Metrics}

\subsection{Functionality of Each Module}

\begin{itemize}
\item \textbf{Extractor}  
Baseline intent extractor that surfaces every candidate \emph{intent–action–tool} triplet from raw marketplace dialogues and product descriptions without filtering or normalization.

\item \textbf{Verifier}  
Gatekeeper that compares each newly extracted item against the current intent space; exact or near-duplicate entries are rejected, preventing redundancy.

\item \textbf{Expert\_guide}  
Domain-expert LLM invoked in a few-shot setting to re-label or re-categorize intents according to predefined taxonomic rules and canonical examples, ensuring semantic consistency across the hierarchy.

\item \textbf{Maintainer}  
Post-processing aggregator that clusters semantically similar intents (via embedding similarity and synonym lists) and collapses redundant nodes, yielding a compact, non-redundant hierarchy while preserving coverage.

\end{itemize}

\subsubsection*{Generation-Quality Metrics}

\paragraph{Coverage \& Duplicate Ratio.}  
We evaluate the mined hierarchy on a 10k marketplace dialogues:

\paragraph{Variables}  
\begin{itemize}
  \item $G$ — number of ground-truth intents in the held-out dialogue set  
  \item $M$ — number of intents our tools successfully match to at least one ground-truth intent  
  \item $T$ — total number of intents we initially extract (before deduplication)  
  \item $U$ — number of unique intents left after removing duplicates
\end{itemize}

\paragraph{Formulas}  
\begin{itemize}
  \item Coverage = $\displaystyle \frac{M}{G}$  
  \item Duplicate Ratio = $\displaystyle \frac{U}{T}$
\end{itemize}

In practice, Coverage > 95 \%, so the smaller the Duplicate Ratio (i.e., the fewer unique intents we keep), the cleaner and higher-quality the final intent space.

\begin{figure}[htbp]
  \centering
  \begin{subfigure}[t]{0.5\textwidth}
    \centering
    \includegraphics[width=\linewidth]{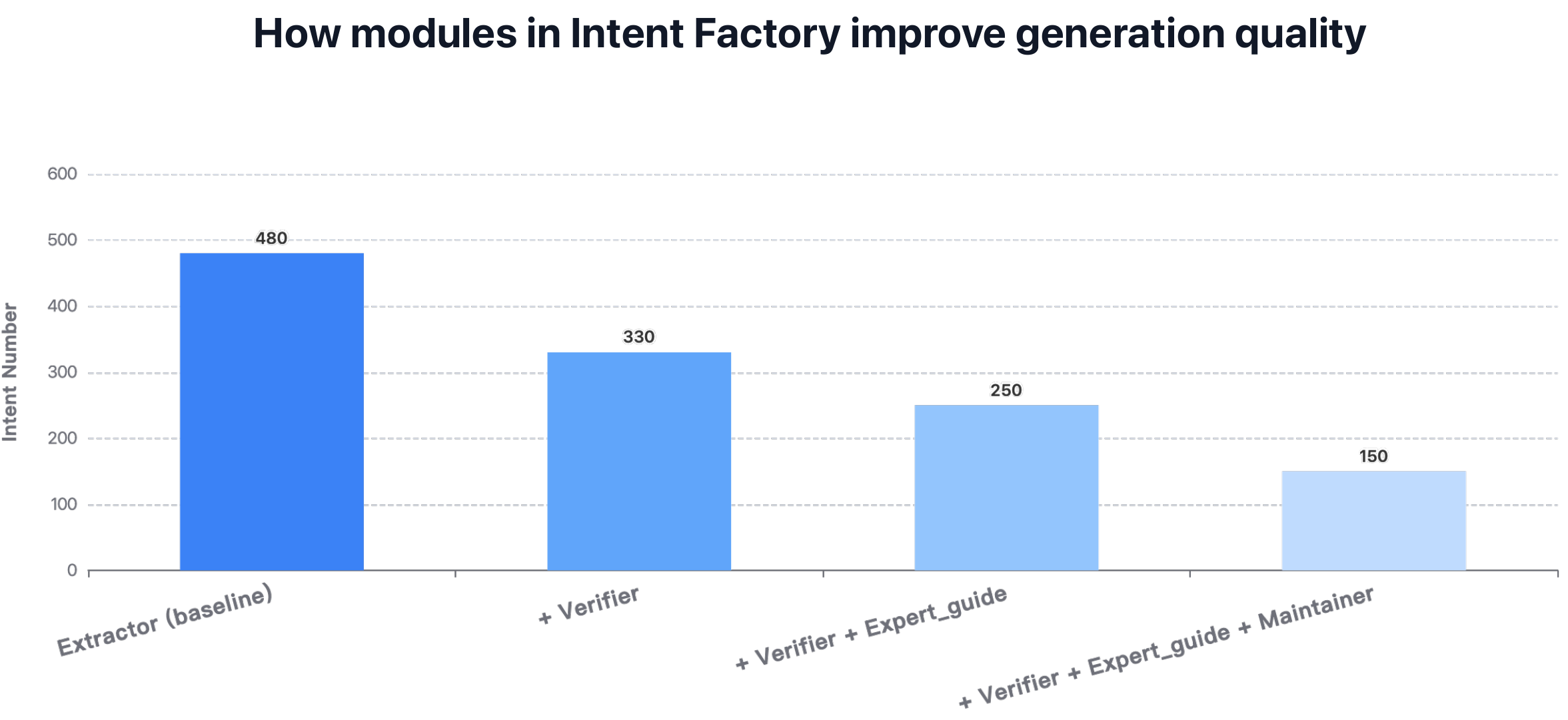}
    \label{fig:ifm}
  \end{subfigure}\hfill
  \begin{subfigure}[t]{0.5\textwidth}
    \centering
    \includegraphics[width=\linewidth]{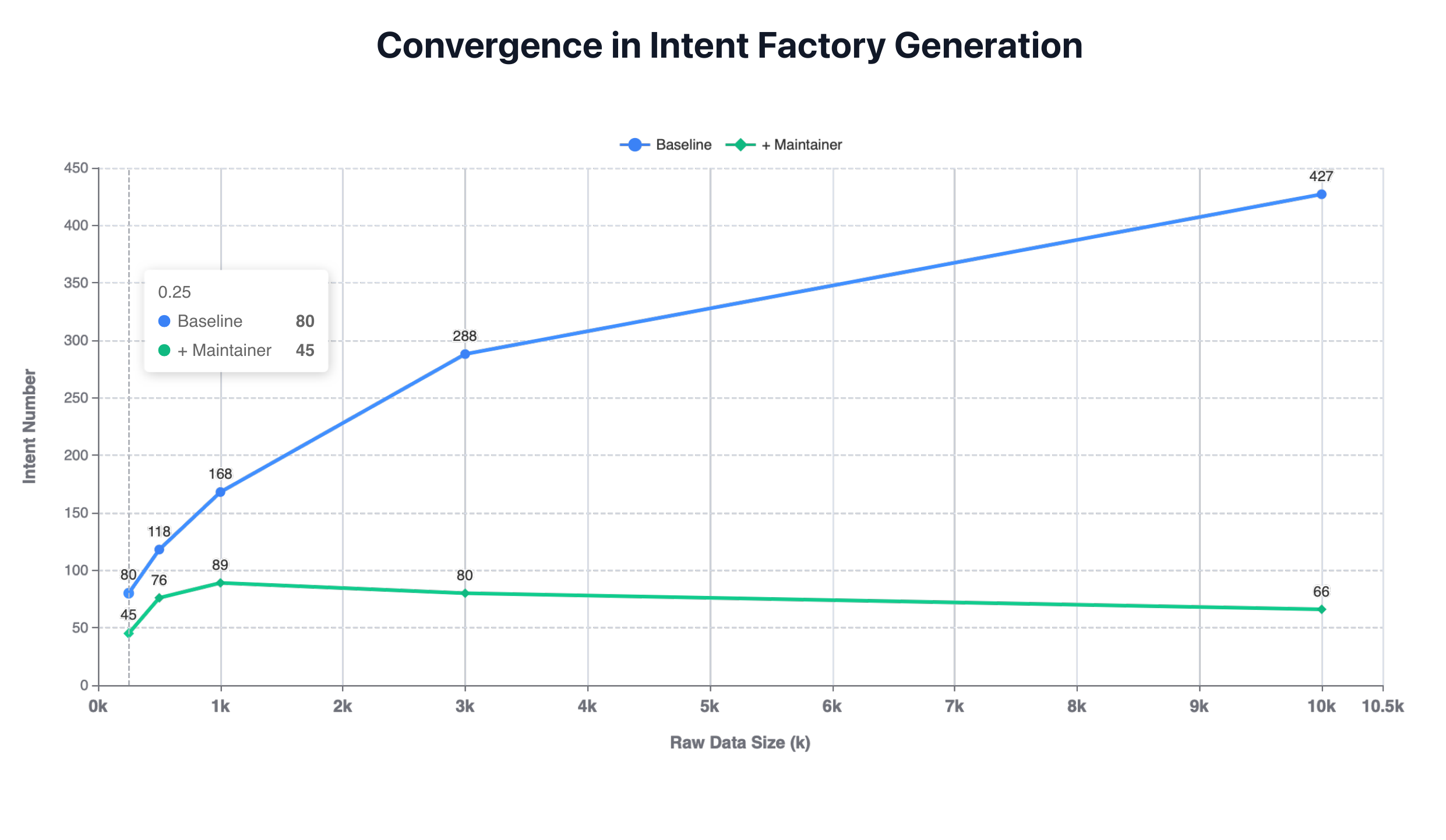}
    \label{fig:ifc}
  \end{subfigure}
  \caption{left: Effect of progressively adding modules refine intent number; right: Convergence behaviour of intent space size as raw data increases}
  \label{fig:two-side-by-side}
\end{figure}

\paragraph{Refinement Curve.}  
Figure~\ref{fig:two-side-by-side} (left) shows how each module progressively reduces the intent count while maintaining coverage above 95 \%. The right-hand plot confirms that the final intent space size converges as raw data increases, indicating bounded growth and stable quality.

\begin{NewBox}*[hb]{box:extractor_prompt}
{The Prompt of extractor}
You will use the Hierarchical Intent Decomposition (HID) framework to analyze the data. This framework is used to extract and structure intents from dialogue data, forming a tree-like hierarchical structure. HID decomposes intents into three orthogonal levels to ensure atomicity, orthogonality, and scalability:

!!! Note, you are an advanced text understanding master and intent recognition expert with rich knowledge. Please fully understand the text and provide intents, not limited to the examples I give.

- Intent (Root): The coarse-grained overall goal of the dialogue 

- Action (Level 1 Branches): Mutually exclusive mid-level stages or categories, with no overlap. These are orthogonal, and if needed, can be expanded through sub-branches.

- Tool (Leaves): Fine-grained atomic operations with a single intent and callable, with parameters (e.g., \{'name': 'API\_QueryPrice', 'description': 'Query item price', 
'returns': \{'price': \{'type': 'number'\}\}\} ). Tools have no child nodes to maintain atomicity.

When processing input data (e.g., raw dialogue JSON with context, history, and features), follow these steps:

1. Extract the root-level Intent based on the overall goal.

2. Decompose into orthogonal Actions, selecting from a predefined set or expanding if necessary (ensuring no overlap).

3. Generate atomic Tools for each Action in JSON object format, including 'name', 'description', and 'returns' (including type/enum where appropriate).

4. For Tool generation, abstract and extract elements from the original text as much as possible, without needing to be very specific.

Ensure the output is consistent, atomic (each Tool has a single purpose), orthogonal (no category overlap), and extensible (if data introduces new intents, suggest new Actions/Tools without violating rules). For the given input, generate the HID decomposition result.

I provide a dialogue segment, where "| buyer |" indicates the buyer's real person role, "| seller |" indicates the seller's real person role, "| bot |" indicates the bot role.
Please generate the HID decomposition result for the "| buyer |" i.e., the buyer's role utterances in the dialogue content.

Output format:
Please output in JSON format, including the keys, in English:

- Intent: string only

- Action: string only

- Tool: JSON format, including attributes like name, 

description, returns, etc., name strictly requires English output, others no requirements
\end{NewBox}

\begin{NewBox}*[hb]{box:verifier_prompt}
{The Prompt of verifier}
You will be responsible for checking whether the newly added Intent-Action-Tool conflicts with the existing action space. Assume that the new Intent-Action-Tool itself is valid (atomicity, clarity, etc., all meet requirements), your only task is to check whether it conflicts with the Intent-Action-Tool in the existing space, including category overlap (non-orthogonal) or functional similarity (duplication).

Each Intent-Action-Tool is a triplet, with "Intent", "Action", "Tool" three key values respectively.

In the action space, all Intent-Action-Tools are organized into three levels: Intent --> Action --> Tool

First, find the corresponding Tools information based on Intent and Action information, and check if the target Tool conflicts with existing Tools.

Verification principles (focusing on conflict detection):
- Each Tool is represented in JSON format, including name, description, parameters, etc., need to compare each piece of information one by one
- Read and understand name, description information to judge whether there is duplication or conflict
- Check if the new Action overlaps with existing Actions (e.g., if existing has 'bargaining', new 'price negotiation' overlaps).
- Check if the new Tool's function is similar to existing Tools, you can assist judgment by checking description and parameters (e.g., if existing has 'API\_QueryPrice', new 'API\_GetItemCost' is functionally duplicate).
- Ensure orthogonality: new items should not cross or copy existing categories.
- If there is no conflict at all, accept; otherwise reject.

Verification steps:
1. Compare the new Intent-Action-Tool with the existing space.
2. Output only one of the following two (strictly follow the format, no additional explanation):
   - "1: No conflict, accept new API"
   - "2: Has conflict, reject"

Input example: Existing space: {'Intent': 'Facilitate transaction', 'Actions': [{'name': 'Information query', 'Tools': [{'name': 'API\_QueryPrice'}]}]}; New item: {'Action': 'Bargaining', 'Tool': {'name': 'API\_ProposeCounteroffer'}}.
Output requirements:
Please output in JSON format, including the following keys:
- status: 1 means no conflict, 2 means has conflict

\end{NewBox}

\subsection{Prompts}
The prompt of extractor is shown in Prompt~\ref{box:extractor_prompt}, and the prompt of verifier is shown in Prompt~\ref{box:verifier_prompt}.

\subsection{An example of intent-tool-action}
There is an example of intent-tool-action shown in Box~\ref{box:intent-tool-action-example}

\begin{NewBox}*[hb]{box:intent-tool-action-example}
{intent-tool-action example}
"Inquire\_Product\_Details": \{
      "Request\_Specification": \{
        "API\_QueryProductSpec": \{
          "description": "Query specific technical parameters of the product, such as power, voltage, or model specifications",
          "parameters": \{
            "spec": \{
              "type": "string",
              "description": "Technical specification requested by the buyer"
            \}
          \}
        \}
      \},
      "Request\_Visual\_Info": \{
        "API\_QueryProductVisualDetail": \{
          "description": "Query specific visual characteristics of the product, especially rear design or presence of elements like numbers",
          "parameters": \{
            "has\_number\_on\_back": \{
              "type": "boolean",
              "description": "Indicates whether the back of the product has a number"
            \},
            "visual\_description": \{
              "type": "string",
              "description": "Textual description of the back appearance"
            \}
          \}
        \}
      \},
\end{NewBox}

\section{Problem Weaver: Detailed Definition}
\label{appendix:problem-weaver}

The \textbf{Problem Weaver} is responsible for transforming abstract entries in the API Pool into concrete, multi-turn bargaining tasks, using publicly available product metadata (\textit{e.g.}, item title, description, listing price, and hierarchical category names) combined with the benchmark's intent pool.

\paragraph{Inputs}
Formally, the module takes as input:  
(1) a product information table
\(\mathcal{P} = \{P_1, P_2, \dots, P_{|\mathcal{P}|}\}\),
where each \(P_k\) contains textual attributes of a real item;  
(2) an intent pool
\(\mathcal{I} = \{I_1, I_2, \dots, I_{|\mathcal{I}|}\}\) produced by the Intent Factory; and  
(3) a set of prompt templates \(\mathcal{T}\) for dialogue generation.

\paragraph{Process}
The Problem Weaver proceeds as follows:
\begin{enumerate}
    \item \textbf{Sampling:} Select a target product \(P_k \in \mathcal{P}\) and an ordered intent sequence \((I_{a_1}, I_{a_2}, \dots, I_{a_m})\) from \(\mathcal{I}\).
    \item \textbf{Verification:} Ensure that the sampled intent sequence is plausible given the product metadata and domain rules.
    \item \textbf{Generation:} Use a prompt template \(T \in \mathcal{T}\) to elicit an LLM-generated buyer utterance for each turn, grounding content in the product attributes.
    \item \textbf{Annotation:} Pair each generated utterance with its corresponding system prompt and gold-standard intent label(s).
    \item \textbf{Structuring:} Aggregate the annotated turns into a structured JSON dialogue object that contains:
        \begin{itemize}
            \item The full scenario script (\emph{dialogue text})
            \item Ground-truth intent labels
            \item The per-turn choice space of candidate intents
        \end{itemize}
\end{enumerate}

\paragraph{Output}
This process yields a synthetic dataset \[
\mathcal{D} = \{D_1, D_2, \dots, D_{|\mathcal{D}|}\},
\]
where each \(D_j\) contains task metadata, scripted buyer–seller dialogues, and per-turn annotations suitable for benchmarking LLM intent recognition in bargaining scenarios.

\subsection{Prompts}
The prompt of problem weaver is shown in Prompt~\ref{box:problem_weaver_prompts}
\begin{NewBox}*[hb]{box:problem_weaver_prompts}
{The Prompt of problem weaver}
Task description
You are a master script-to-task writer.  
Your ONLY inputs are:

1. product\_info - a short text containing the item description, price, and category.
2. ground\_truth\_action - an ordered list of API calls that the buyer must eventually issue.

Notes: Product info are match with format: item\_desc, item\_price, channel\_cate\_level1\_name, channel\_cate\_level2\_name, channel\_cate\_level3\_name,channel\_cate\_level4\_name

Your job is first generate  {buyer\_question} in English that naturally triggers the ground\_truth\_action.  
Feel free to add a plausible personal context so the question looks realistic.

Rules  
- Keep the question under 40 words.  
- Mention only the **first** API in the Ground Truth list; do not reveal the rest.  
- Translate any Chinese terms in product\_info into natural English.  
- Do not quote the API names literally; phrase the concern in everyday language.

---
Case

Sample Input  
product\_info:  
Sam's Club Elsa Princess Dress, size 140. Worn once for photos—like new. ¥58. Kids' Apparel > Dresses > Princess Dresses. 
[API\_CheckHeightFit, API\_QueryShippingPolicy, API\_CalculateOfferPrice]

Above action refers to:
 "API\_CheckHeightFit": \{
          "description": "Check if the product (e.g., bicycle) is physically suitable for the buyer based on their height or body measurements",
          "parameters": \{
            "fit\_result": \{ "type": "string", "enum": ["suitable", "too\_small", "too\_large", "uncertain"] \},
            "reason": \{ "type": "string" \}
          \}
        \}
      \}
    \},

    "Inquire\_Shipping\_Logistics": \{
      "Check\_Shipping\_Policy": \{
        "API\_QueryShippingPolicy": \{
          "description": "Query whether the item is eligible for free shipping based on product details and seller settings",
          "parameters": \{
            "free\_shipping": \{ "type": "boolean", "description": "Indicates if the item is eligible for free shipping" \},
            "shipping\_fee": \{ "type": "number", "description": "The shipping cost if not free, in yuan" \},
            "shipping\_notes": \{ "type": "string", "description": "Additional notes about shipping" \}
          \}
        \},

        "API\_CalculateOfferPrice": \{
          "description": "Calculate a reasonable offer price based on item's marked price, bottom price, and negotiation stage",
          "parameters": \{
            "offered\_price": \{ "type": "number" \},
            "shipping\_included": \{ "type": "boolean" \}
          \}
        \},

Sample "buyer\_question" output:
“My daughter is 135 cm—will the size 140 be too big for her? Could you do 50 yuan with free shipping?”

---
Inputs

product\_info: {product\_info}

ground\_truth\_action: {ground\_truth\_action}

---

Output Format

Your output should strictly follow the format. Otherwise, a cute kitty will starve for a dineer.

Ignore the parameters for now.

Please output in JSON format, including following keys:
"buyer\_question": string, a single-turn buyer question in English, in natural language.

\end{NewBox}

\section{Discussion on Framework Advantages}
\label{appendix:framework-advantage}

Taken together, the \textbf{Intent Factory}, \textbf{Problem Weaver}, and \textbf{Evaluation Center} form an integrated pipeline for constructing and administering controlled bargaining evaluations. The bottom-up design ensures that every evaluation instance originates from realistic scenario data, is framed by a well-defined \emph{intent–action–tool} hierarchy, and is paired with explicit turn-level ground truth. 

Unlike benchmarks that stage end-to-end negotiation matches and judge performance by win–loss outcomes, our framework isolates the specific capability of \emph{understanding} bargaining context: models are asked to infer buyer intent from dialogue history and structured choice spaces, rather than to simply generate plausible conversation turns. This design exploits the asymmetry between authoring and solving — models can readily produce convincing multi-turn interactions when given target intents, yet often fail to reliably recover those intents from completed exchanges.

By preserving verifiable ground truth at each turn, our method delivers interpretable, reproducible, and fine-grained performance measurements. Furthermore, because it is grounded in general principles of intent extraction, scenario synthesis, and structured evaluation, the approach can be directly adapted to other multi-turn, goal-oriented domains such as diplomatic negotiations, collaborative planning, and multi-party discussions.




\section{Data Preparation Details}
\label{appendix:data-prep-details}

The benchmark dataset is constructed through the integrated pipeline of \textit{Intent Factory}, \textit{Problem Weaver}, and \textit{Evaluation Center}.  

The \textbf{Intent Space} is derived from 10k authentic second-hand marketplace dialogues, focusing on extracting and aggregating buyer intents. We employ the advanced \texttt{qwen-plus-latest} model to perform large-scale extraction and refinement, consuming approximately 400M tokens; detailed prompting strategies are provided in the Appendix. The resulting structure contains 17 intents, 39 actions, and 65 tools (tools is the most granular level of intent hierarchical tree).  

\begin{figure}[htbp]
    \centering
    \includegraphics[width=0.65\textwidth]{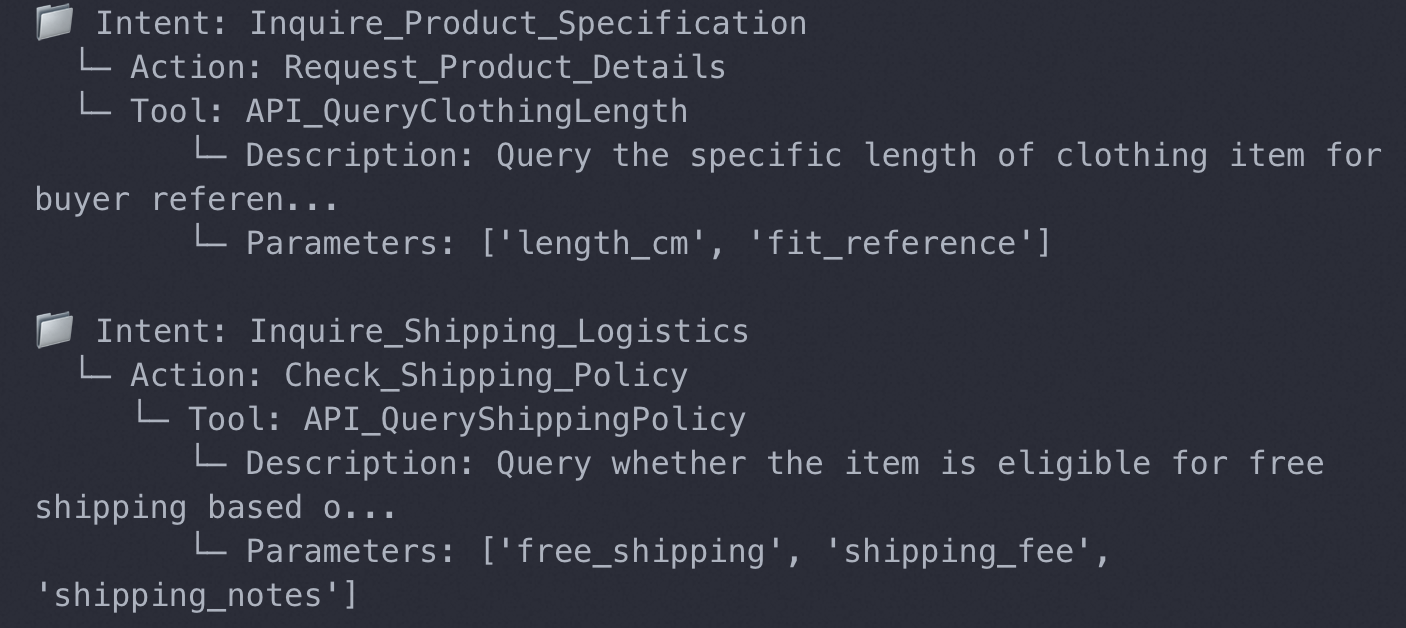}
    \caption{Intent 66: finalized version of hierarchical intent space.}
    \label{fig:intent66}
\end{figure}

\begin{figure}[htbp]
    \centering
    \includegraphics[width=0.65\textwidth]{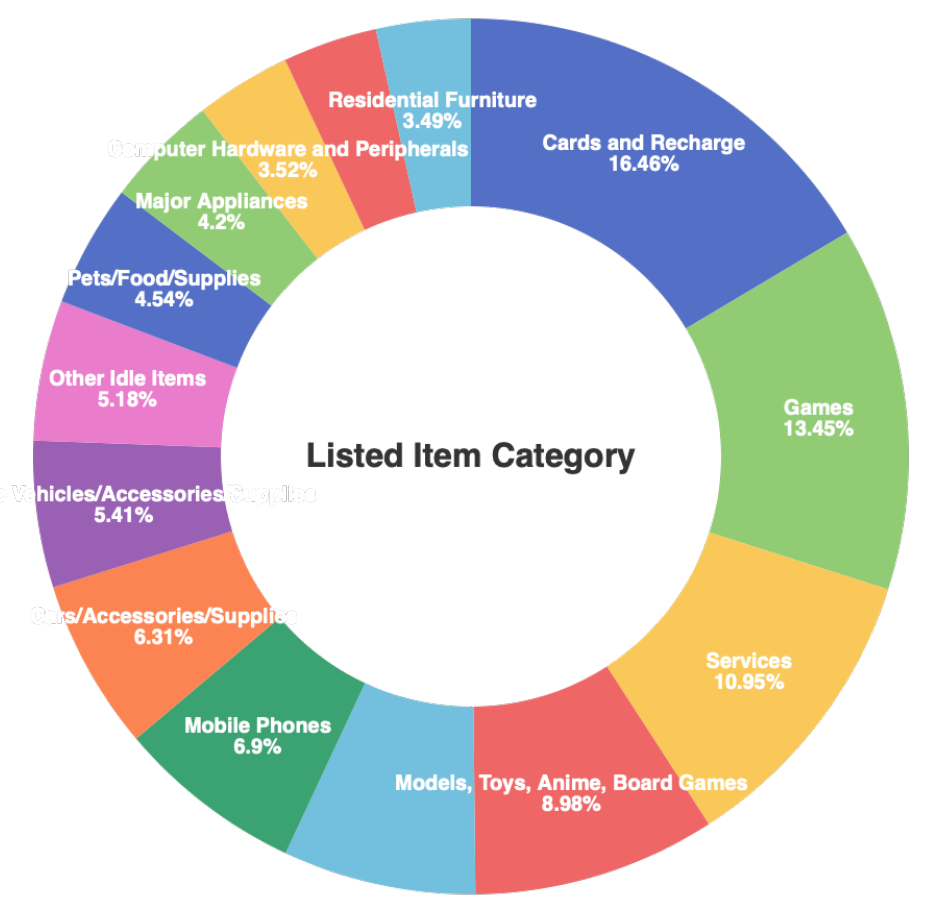}
    \caption{Pie chart of listed item Level 1 categories.}
    \label{fig:category-proportion}
\end{figure}

\begin{table}[ht]
\centering
\caption{Dataset Statistics}
\begin{tabular}{l r}
\toprule
\textbf{Statistic} & \textbf{Value} \\
\midrule
Total Items & 9,892 \\
Average Price & 1,487.17 \\
Unique Level 1 Categories & 85 \\
Unique Level 2 Categories & 700 \\
Unique Level 3 Categories & 1,336 \\
Unique Level 4 Categories & 1,611 \\
\bottomrule
\end{tabular}
\label{tab:dataset-stats-appendix}
\end{table}

\begin{figure}[htbp]
    \centering
    \includegraphics[width=0.65\textwidth]{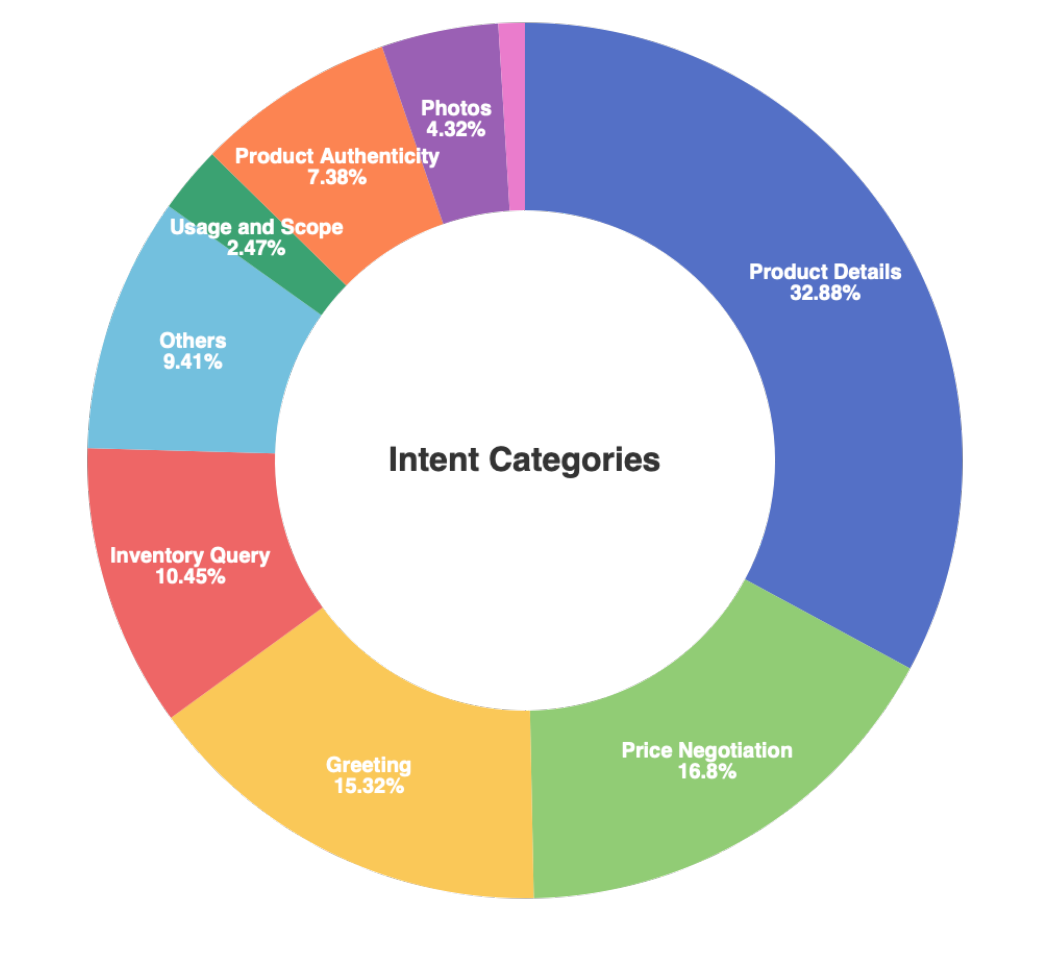}
    \caption{Pie chart of intent space.}
    \label{fig:intent-pie}
\end{figure}

\section{Evaluation Task Sample}
\begin{figure}[htbp]
    \centering
    \includegraphics[width=0.65\textwidth]{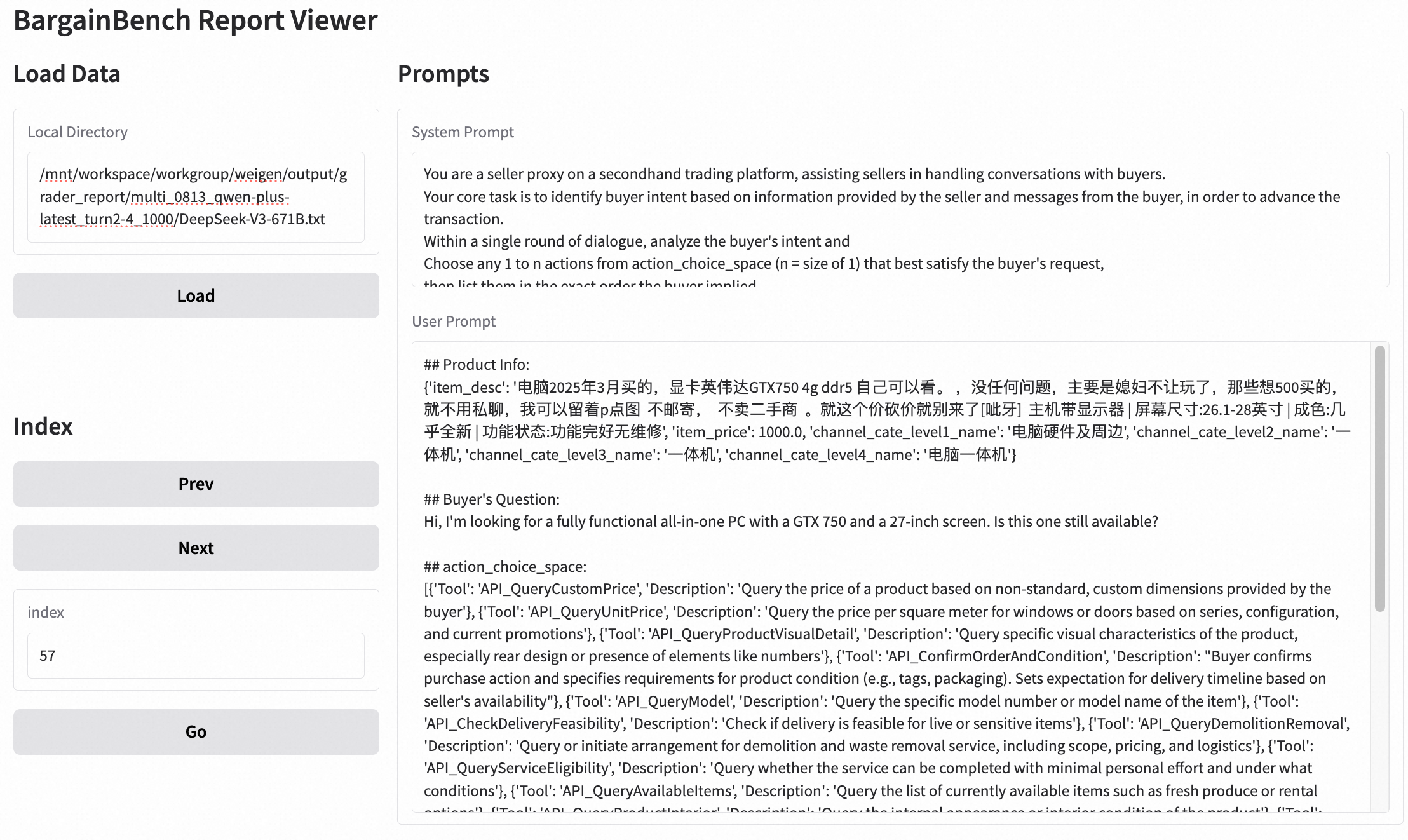}
    \caption{An illustration of evaluation sample}
    \label{fig:task-sample}
\end{figure}

\begin{figure}[htbp]
    \centering
    \includegraphics[width=0.65\textwidth]{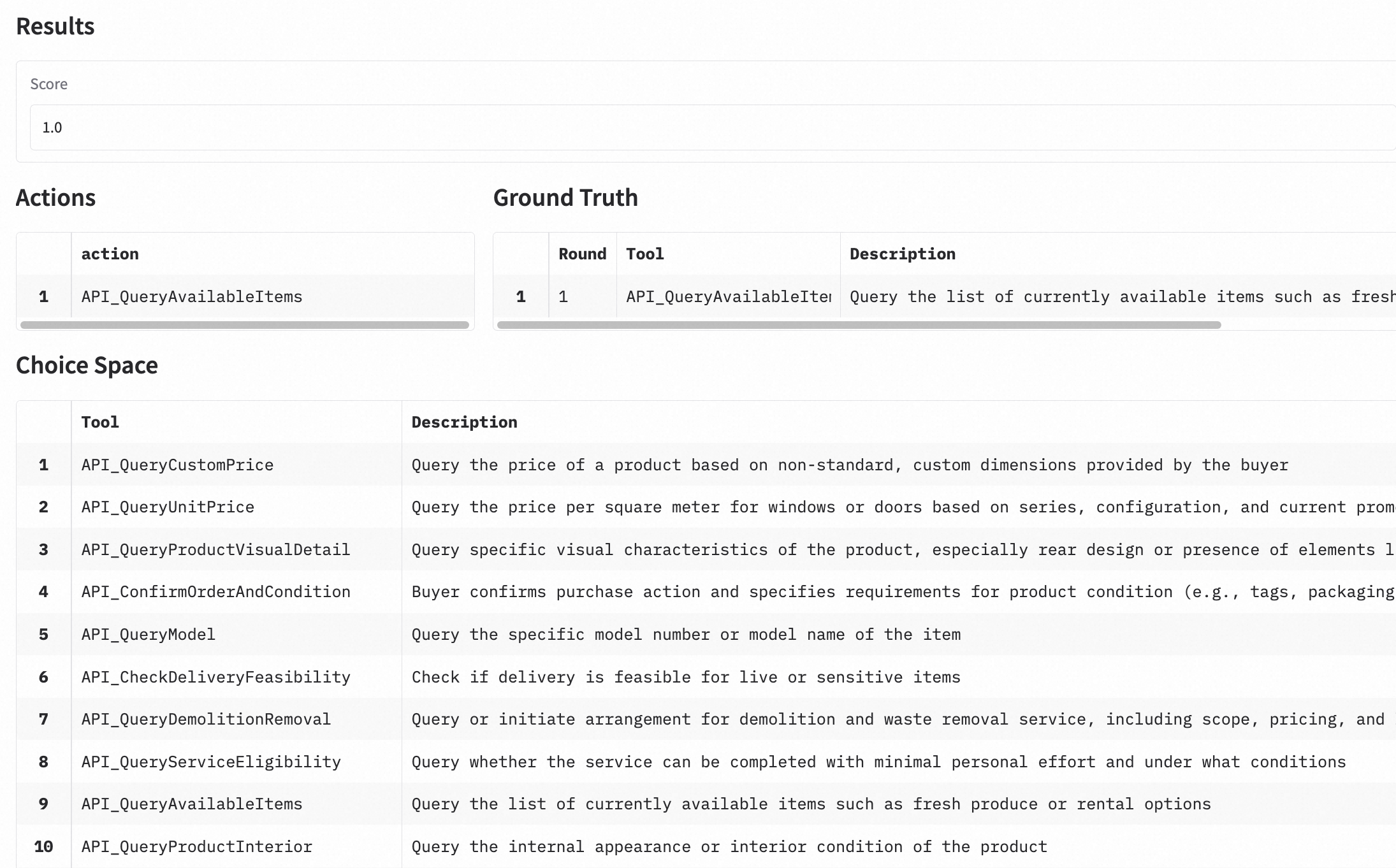}
    \caption{Candidate Intents}
    \label{fig:MC-sample}
\end{figure}

There is a case of evaluation result in Figure~\ref{fig:task-sample} and Figure~\ref{fig:MC-sample}. Task generated by problem weaver, consist of system prompt, product info and context. Candidate model have to choose the best fit intent from a intent space of 20.

\label{appendix:task-sample}

\section{Extended Related Work}
\label{appendix:related}

Here we include the extended discussion of related work. 

\textbf{Multi-Turn Interaction and Negotiation Benchmarks.} 
Early datasets such as DealOrNoDeal \cite{lewis_deal_2017} and CraigslistBargain \cite{he_decoupling_2018} established text-based bargaining protocols. Recent benchmarks move toward interactive settings: Xia et al.\ \cite{xia_measuring_2024} formalize bargaining as an asymmetric incomplete information game, while Davidson et al.\ \cite{davidson_evaluating_2024} evaluate model agency through negotiation games. More applied systems, such as FishBargain \cite{dexin_fishbargain_2025} and debt collection frameworks \cite{wang_debt_2025}, extend bargaining research to real-world domains.

\textbf{Intent Recognition and Tracking in Dialogue.} 
Dialogue State Tracking datasets such as MultiWOZ \cite{budzianowski_multiwoz_2018} address explicit task goals, but bargaining often involves implicit and shifting intents. Guan et al.\ \cite{guan_evaluating_2025} survey methods for multi-turn conversations, emphasizing the challenge of intent tracking. NegotiationToM \cite{chan_negotiationtom_2024} stresses belief and intention modelling, showing that even advanced models struggle to maintain consistent inference across turns.

\textbf{Tool Use in Multi-turn Systems.} 
Benchmarks for tool-augmented dialogue agents emphasize correctness and robustness. $\tau$-Bench \cite{yao_-bench_2024} evaluates tool--agent--user interaction under domain-specific rules. ToolACE \cite{liu_toolace_2025} provides large-scale synthesized function-calling data, while ACEBench \cite{chen_acebench_2025} categorizes tool-use evaluation into multiple scenarios. These works highlight challenges in aligning user intent, action, and tool execution.

\textbf{Distinction from Prior Work.} 
Our work complements these studies by shifting the focus from outcome-based metrics to turn-level buyer intent recognition, explicitly grounded in an \emph{intent–action–tool} hierarchy. Compared to prior multi-turn benchmarks \cite{guan_evaluating_2025, yao_-bench_2024, liu_toolace_2025, chen_acebench_2025}, we emphasize semantic alignment between intent, action, and tool in bargaining, connecting to broader social intelligence evaluations such as SOTOPIA \cite{zhou_sotopia_2024}.

\bibliographystyle{plainnat}

\bibliography{references}

\end{document}